\documentclass[9pt,shortpaper,twoside,web]{ieeecolor}
\usepackage{generic}
\usepackage{cite}
\usepackage{amsmath,amssymb,amsfonts}
\usepackage{algorithmic}
\usepackage{graphicx}
\usepackage[hidelinks]{hyperref}
\usepackage{float}
\usepackage{algorithm}
\usepackage{amsmath}
\usepackage{multirow} 
\usepackage{textcomp}
\def\BibTeX{{\rm B\kern-.05em{\sc i\kern-.025em b}\kern-.08em
    T\kern-.1667em\lower.7ex\hbox{E}\kern-.125emX}}
\begin{document}
\title{Towards Modality-Agnostic Seizure Classification: A Unified Deep Learning Framework for Heterogeneous EEG and Intracranial EEG Signals}
\author{Ramya Sundaram, \IEEEmembership{Member, IEEE},  Muhammad Aslam Jarwar, \IEEEmembership{Senior Member, IEEE}, M Uma , Sunder Ali Khowaja, \IEEEmembership{Senior Member, IEEE} and Qurat-Ul-Ain Mastoi, \IEEEmembership{Senior Member, IEEE}
\thanks{Ramya Sundaram is with School of Computing and Digital Technologies, Sheffield Hallam University, Sheffield and University of Strathclyde, Glasgow, United Kingdom. ramya.sundaram@strath.ac.uk}
\thanks{Muhammad Aslam Jarwar is with  School of Computing and Digital Technologies, Sheffield Hallam University, Sheffield, United Kingdom (corresponding author, e-mail: a.jarwar@shu.ac.uk).}
\thanks{M Uma is with Department of Cmputational Intelligence, SRM Institute of Science and Technology,Kattankulathur-603203, India. umam@srmist.edu.in}
\thanks{Sunder Ali Khowaja is with Faculty of Engineering and Computing,  Dublin City University, Ireland. sunderali.khowaja@dcu.ie }
\thanks{Qurat-Ul-Ain Mastoi is with  Department of Computer Science and Creative Technologies, University of the West of England, Bristol, United Kingdom. Qurat-Ul-Ain.Mastoi@uwe.ac.uk.}
}

\maketitle

\begin{abstract}
Seizure prediction and detection, essential for clinical treatment of epilepsy, remain the crucial task of this neurological disease. EEG and iEEG are widely used to analyze epilepsy due to its capability of capturing discriminative brain activities at high time resolution. However, existing works generally analyze either EEG or iEEG seizure activity separately, where EEG analysis has been well-researched, while more and more works consider iEEG for its higher specificity at the spatial level and signal quality. Currently, most of the existing seizure prediction systems are single-modal in nature, which limits its application to a variety of clinical conditions. Therefore, a common framework to analyze seizures with both EEG and iEEG signals needs to be developed. This paper presents a new framework of multimodal seizure classification integrating the heterogeneous EEG and iEEG recordings from different subjects and sources. A specific framework, called Unified Multimodal Signal Harmonization (UMSH), is developed for harmonizing signals with varied sampling rates and recording conditions. Moreover, to handle the extremely class-imbalanced problem in the data sets, we use a Conditional Generative Adversarial Network (cGAN) based data augmentation technique to generate synthetic samples to balance the class distribution. The proposed framework uses the EEGNet-Vision Transformer (EEGNet-ViT) as the feature extraction and classification network to learn the local discriminative patterns as well as global contextual information in the multimodal seizure data. Experimental results demonstrate the effectiveness of the proposed framework for robust seizure classification, emphasizing the potential of integrated EEG-iEEG multimodal learning in advancing next-generation epilepsy monitoring and clinical decision-support systems.   
\end{abstract}


\section{Introduction}
\label{sec:introduction}
	Epilepsy is a common neurological disorder and the fundamental characteristic feature is the presence of recurrent seizures due to abnormal electrical discharges in the brain. The condition affects more than 50 million people globally, seriously jeopardizing their well-being and quality of life  \cite{anwar2020epileptic}. Early prediction and detection of seizures is crucial to allow timely management of epilepsy. Accurate prediction can provide advance notice to both patient and caretaker, improving their safety. Missed detection or delayed detection of seizures can cause damage, prolonged seizures, hospitalization, and serious long-term implications  \cite{rasheed2020machinelearning}. The research challenge of devising an automatic system for seizure detection continues to be addressed. Electroencephalography (EEG) has become a standard procedure for epilepsy monitoring, which measures the electrical activity of the brain through electrodes placed on the scalp. It is a preferred choice as it is inexpensive, non-invasive and commonly available  \cite{benbadis2020role}. One of the major drawbacks of EEG is its noisy and spatially poor recordings that result in imprecise localization of seizure focus. Intracranial electroencephalography (iEEG) overcomes this limitation as it records from electrodes surgically placed on the brain's surface or within its structures, resulting in greater signal accuracy and spatial resolution. However, due to the invasive nature, increased cost, and limited availability it is not suitable for regular monitoring \cite{kolodziej2023implementation}.
With the evolution of artificial intelligence, various automated seizure detection and prediction systems have been built using both EEG and iEEG signals. In conventional machine learning techniques, it usually consists of manual feature extraction methods, including time-domain, frequency-domain, and time-frequency methods, then using classifiers like SVM, Random Forests, and k-NN \cite{hernandez2018detecting}. Recently, deep learning-based frameworks, namely CNNs, RNNs, LSTMs, and transformer-based models, have been shown better performance in this task as these models are able to learn useful discriminative representations directly from neural signals without the help of manually designed features \cite{prasanna2026multiseizure}. Both EEG \cite{kode2024seizuredetection} and iEEG \cite{sun2022continuous} studies achieve good classification accuracy, showing the potential of AI in seizure analysis.
Existing seizure detection frameworks are promising in their performance \cite{chung2024singlechannel} and validation analysis \cite{esmaeilpour2024deeplearning}. Current seizure prediction methods have shown encouraging prediction ability, but several models are commonly tailored to a certain signal modality or a fixed recording environment, reducing their usefulness for diverse clinical situations. In routine seizure monitoring, patients may undergo non-invasive scalp EEG monitoring during standard or prolonged observation; however, when more localized neurological information is required, intracranial EEG (iEEG) may be obtained during specialized invasive assessment. This complementary monitoring conditions lead to prediction frameworks that are flexible to heterogeneous neural signals and not limited to a single acquisition modality. Inspired by the outlined demand, we propose a seizure prediction framework that incorporates various EEG/iEEG to accommodate heterogeneous conditions. The overall framework is designed as a prospective prediction unit for the real-time clinical monitoring envisioned for the 6G-IoMT scenarios, where different physiological signals are transmitted for assessing seizure risk and timely warning of doctors.
This paper proposes to build a multimodal unified seizure dataset by merging EEG and iEEG recorded from diverse datasets and varied subjects. The core goal is to devise a common space for representing the seizure information extracted from both modalities within a unified learning framework. For this purpose, a customized pre-processing framework, UMSH (Unified Multimodal Signal Harmonization) is proposed. UMSH works on homogenizing the heterogeneous signals by synchronizing sampling frequencies, normalizing, segmenting, and standardizing data in order to combine EEG and iEEG into one multimodal dataset, on which the seizure classification framework is to be applied to allow learning with the benefit of both modality-specific knowledge and heterogeneous information.
A fundamental issue in the proposed multimodal architecture emerges from the integration of EEG and iEEG datasets gathered from diverse sources, specifically the CHB-MIT \cite{guttag2010chbmit} EEG dataset and a Kaggle-based iEEG dataset \cite{nejedly2020multicenter}. Due to disparities in class distribution and imbalances in the quantity of samples available per seizure category, the combined multimodal dataset has a severe class imbalance. Class imbalance can create models that overfit to the majority classes, resulting in poor classification of the minority classes. To address this imbalance, a Conditional Generative Adversarial Network (cGAN)-based data augmentation approach is applied to create a realistic set of synthetic samples of under-sampled classes, improving the balance and variety within the dataset. Once the data is augmented, a hybrid EEGNet-Vision Transformer (EEGNet-ViT) model is used to classify the seizures. The EEGNet model is effective at extracting local space-temporal discriminative features from the combined signals, while the Vision Transformer excels at learning long-range dependencies and global contextual interactions from the data. Combined, the strengths of these models are used to learn comprehensive multimodal representations to classify seizures effectively.

This study focuses on the design, development, and validation of a generalized seizure prediction system based on deep learning, specifically using EEGNet and Transformer-based models. The research investigates the role of spatial-temporal attention and data augmentation (via CGANs) in improving seizure classification performance across heterogeneous EEG datasets.

The major contributions of this work are summarized as follows:
\begin{itemize}
    \item A novel multimodal seizure classification framework integrating EEG and iEEG data from independent datasets is proposed.
    \item A Unified Multimodal Signal Harmonization (UMSH) pre-processing framework is developed to address modality heterogeneity and varying sampling frequencies. 
    \item A cGAN-based augmentation strategy is employed to mitigate class imbalance and improve data diversity. 
    \item A hybrid EEGNet-Vision Transformer architecture is designed to effectively learn local and global representations for seizure classification. 
    \item Comprehensive experiments demonstrate the effectiveness of multimodal EEG-iEEG learning for robust epilepsy analysis.
\end{itemize}

\section{Related Works}
The automatic detection and classification of seizures has attracted a lot of attention in recent years. Many researchers used different machine learning and deep learning approaches to classify brain signals related to seizures. The most part of the research used EEG signals for seizure analysis, while the amount of research related to iEEG was relatively small. To gain knowledge about these methods, it is necessary to understand the recent trends and limitations. Hence, in this section, the existing work related to automated seizure detection and classification using both EEG and iEEG is reviewed, especially the machine learning and deep learning approaches utilized.
\subsection{EEG-Based Seizure Detection and Classification}
EEG-based seizure detection and classification have been widely studied using various machine learning and deep learning techniques \cite{kunekar2024detection}. Among the publicly available datasets, the CHB-MIT \cite{guttag2010chbmit} and University of Bonn datasets \cite{andrzejak2001indications} are the most commonly used due to their availability and well-annotated seizure recordings. Additionally, the Temple University Hospital EEG Seizure Corpus (TUSZ) \cite{shah2018tuh}, one of the largest clinical EEG datasets, is frequently utilized for large-scale and cross-subject seizure classification studies. In machine learning-based seizure detection and classification studies, handcrafted features extracted from EEG signals are commonly employed for model development. Researchers have extensively investigated time-domain, frequency-domain, and time–frequency-domain feature extraction techniques \cite{hernandez2018detecting} to capture seizure-related characteristics and improve classification performance. While some studies have focused on single-channel EEG \cite{chung2024singlechannel}  analysis to reduce computational complexity, others have explored channel selection strategies \cite{dokare2025optimized} to identify the most informative EEG channels and enhance classification accuracy. Notably, using GWO-based optimization of features along with hybrid SVM-Fuzzy classification enhanced the performance for the diagnosis of seizures \cite{badr2025svmfuzzy}, and the use of Mutual information and Random Forest-based channel selection using DWT and SVM allowed the detection of seizures in a computationally efficient way \cite{dokare2025optimized}. Additionally, CNN-enhanced feature learning and epileptic state classification were combined with dynamic channel selection, proving the efficacy of optimal channel selection techniques in seizure detection \cite{awan2026adaptive}.
Many studies have developed hybrid frameworks that use feature selection \cite{ghosh2026seizure}, feature fusion \cite{jayanthi2026quantum}, and sophisticated classification methods to enhance seizure detection performance \cite{heba2026epilepsyformer}. Particularly, a CNN-LSTM-based \cite{kashefi2025cnn_lstm} model was created to extract spatial and temporal information from DWT-derived EEG sub-bands, with excellent detection performance across the TUSZ, Bonn, and CHB-MIT datasets. A multiple-input deep feature learning network (MDFLN) \cite{sun2024mdfln} learning both the time and the time-frequency domain features was presented, and BLSTM was employed for seizure classification, where the proposed network performance surpassed that achieved using a single view feature representation. Moreover, EfficientNet-B0-based \cite{saadoon2025efficientnet} CNN combined with the ensemble SVM voting method trained with STFT and channel correlation features demonstrates an effective approach for reliable prediction, achieving comparable accuracy to the existing research with CHB-MIT data. Besides these binary detection approaches, several works have investigated multiclass seizure classification to distinguish among different seizure types and thus improve the clinical utility of the TUSZ dataset \cite{chen2025multiband}.

\subsection{iEEG-Based Seizure Detection and Classification}
Besides methods based on scalp EEG, a number of research groups have explored iEEG signals for the purpose of seizure detection and prediction due to their better signal quality and spatial resolution. Recent surveys indicated that machine learning and deep learning methods for the detection of epileptogenic zones using iEEG signals are receiving wide attention \cite{daida2025ai}. In order to achieve the improvement of seizure detection accuracy on long-term iEEG data, Laelaps (a hyperdimensional computing-based approach) was proposed \cite{burrello2019laelaps}, and a high detection rate with the lowest energy and no false alarms was achieved. To develop a patient-independent seizure detection system, transformer-based and multi-task learning models have been studied \cite{sun2025multitask}, and the proposed models showed high generalization ability on the patients and datasets. For the issue of low-resource devices, a lightweight CNN model with an attention mechanism has been designed for seizure detection in iEEG signals \cite{si2023lightweightcnn}. For epilepsy state classification using iEEG, CNN, CNN-SVM, and hybrid CNN-based models have been developed \cite{abderrahim2024ieeg}. Moreover, a convolutional-transformer-based model has been studied for effectively utilizing inter-channel information and to obtain an interpretable model in seizure detection based on continuous long-term iEEG recordings \cite{sun2022continuous}.
According to the current literature, most seizure detection and prediction investigations have used either scalp EEG or iEEG data independently. Despite substantial advances in both fields, no study has extensively studied a single system capable of managing both EEG and iEEG inputs concurrently. This restriction is mostly due to fundamental differences between EEG and iEEG recordings, such as variances in signal properties, electrode topologies, spatial resolution, noise levels, and sampling frequencies. Hence, it is difficult to create a generalized model that is efficient to learn both modalities of signal. This research gap is addressed in this study by developing a unified seizure classification framework for the EEG and iEEG signals. The proposed approach attempts to increase the generalizability and flexibility of seizure detection systems, allowing a single model to function across several recording modalities and assisting in the creation of more flexible epilepsy monitoring systems.

\section{PROPOSED METHODOLOGY}
\begin{figure}[t]
\centering
\includegraphics[width=0.95\columnwidth]{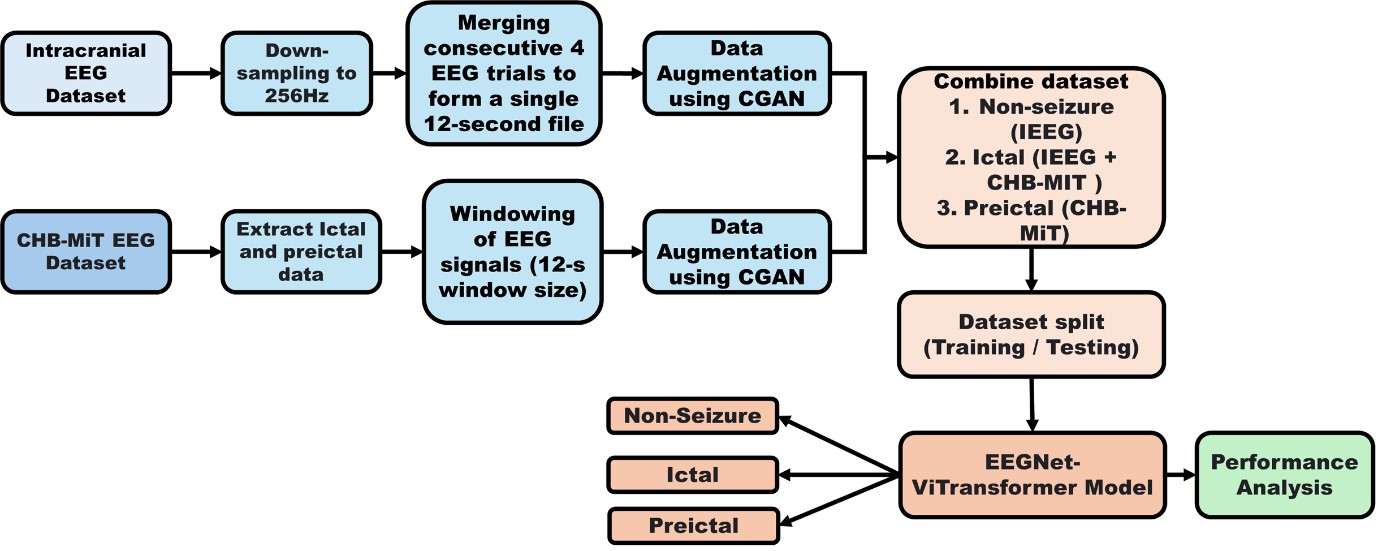}
\caption{General block diagram of the proposed methodology.}
\label{fig:figure1}
\end{figure}

In this work, a single framework for seizure classification using both scalp EEG and iEEG signals is presented. The proposed method integrates the EEG and iEEG datasets and forms a generalized seizure classification model that can accommodate both recording modalities. Since EEG and iEEG signals have distinct differences in their characteristics, channel setup, and sampling frequencies, the proposed methodology first uses a specialized pre-processing technique, namely, Unified Multimodal Signal Harmonization (UMSH), to convert EEG and iEEG signals into similar representations for downstream tasks. The merger of these two datasets also forms a strongly imbalanced classification problem, which is dealt with using an adequate balancing method. After preprocessing and balancing the data, features are extracted and classified using the hybrid EEGNet and ViT models to exploit discriminative patterns relevant to seizures. The elaboration of each step of the proposed method is given in the following sub-sections. The entire structure of the proposed methodology is shown in Figure \ref{fig:figure1}.

\begin{algorithm}[H]
\caption{Unified Multimodal Signal Harmonization (UMSH)}
\label{alg:umsh}
\small
\textbf{Input:} Multicentre iEEG Dataset, CHB-MIT Dataset\\
\textbf{Output:} Harmonized Unified EEG--iEEG Dataset
\begin{enumerate}\itemsep 2pt
  \item Load $\mathrm{Dataset}_{\mathrm{Fnusa}}$ and $\mathrm{Dataset}_{\mathrm{Mayo}}$.
  \item Resample iEEG from 5000\,Hz to 256\,Hz (15000 to 768 samples).
  \item Concatenate 4 consecutive samples to obtain 3072 samples:\\
        $X_i = \mathrm{Concat}(x_i, x_{i+1}, x_{i+2}, x_{i+3})$
  \item Store ictal and non-seizure samples.
  \item Load CHB-MIT EEG recordings.
  \item Compute channel average of Fp1--F3, P7--O1 and P8--O2:\\
        $X_{\mathrm{avg}}(n) = \dfrac{X_{\mathrm{Fp1\text{-}F3}}(n) + X_{\mathrm{P7\text{-}O1}}(n) + X_{\mathrm{P8\text{-}O2}}(n)}{3}$
  \item Extract ictal segments.
  \item Extract preictal segments (10-min before seizure).
  \item Segment recordings into 12-sec windows.
  \item Generate EEG samples of 3072 points.
  \item Perform data standardization.
  \item Combine EEG and iEEG into unified dataset.
\end{enumerate}
\end{algorithm}

\begin{figure}[t]
\centering
\includegraphics[width=0.95\columnwidth]{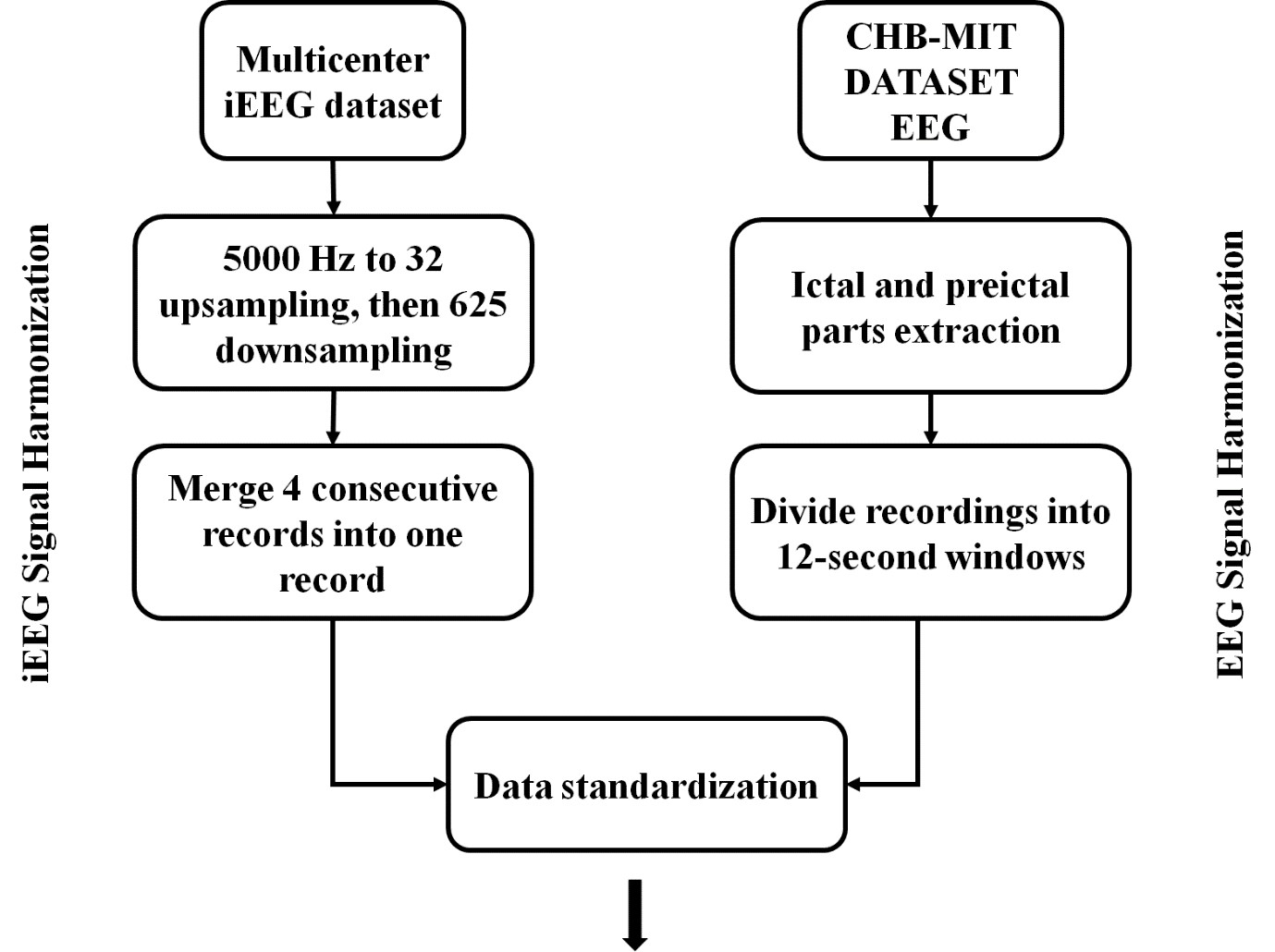}
\caption{Proposed Multimodel dataset creation (Unified Multimodel Signal Harmonization.}
\label{fig:figure2}
\end{figure}

\begin{figure}[t]
\centering
\includegraphics[width=0.95\columnwidth]{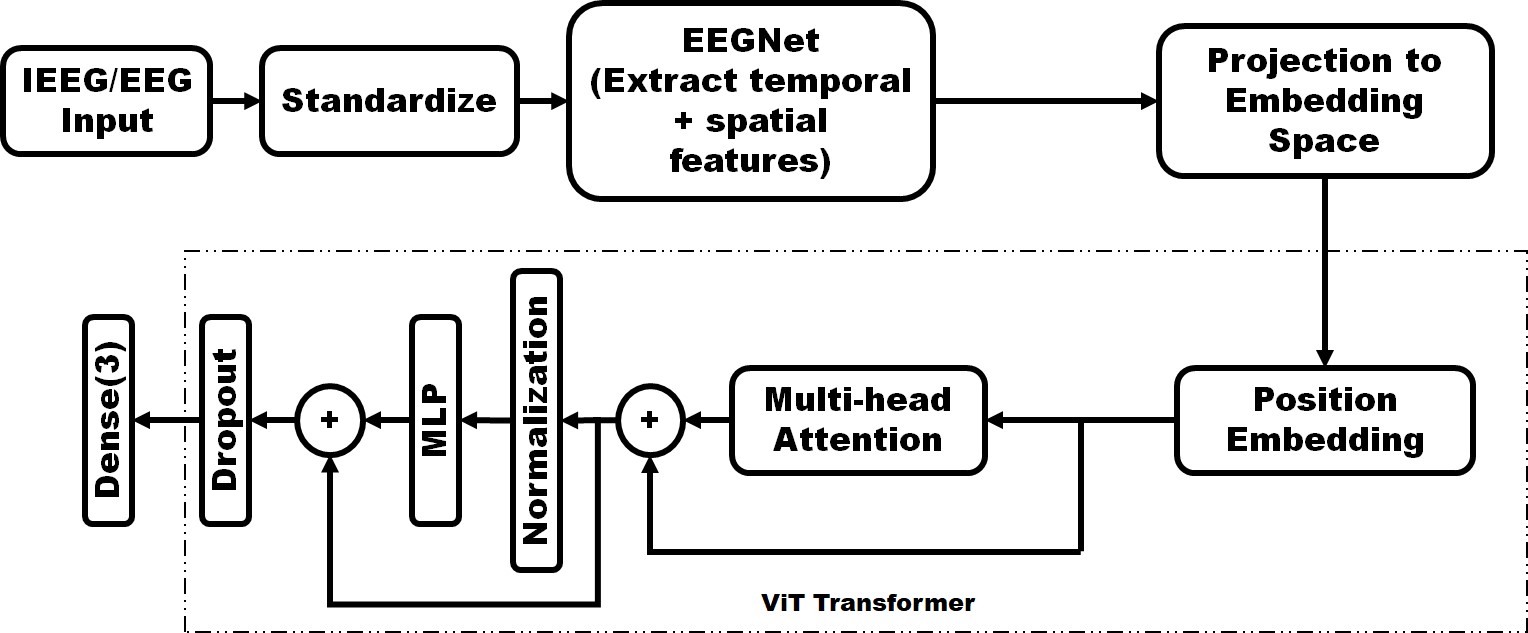}
\caption{General block diagram of the EEGNet-ViT Architecture for multimodal classification.}
\label{fig:figure3}
\end{figure}
\subsection{Dataset Creation}
In this paper, we combine both EEG (scalp and iEEG) and form a single dataset to create a generalized seizure classification method. This was performed using the CHB-MIT EEG database and the Mayo Clinic iEEG dataset from Kaggle.

\begin{table}[H]
\centering
\caption{Dataset Details of the iEEG Dataset}
\label{tab:ieeg_dataset_details}
\resizebox{\columnwidth}{!}{
\begin{tabular}{|l|c|c|c|c|}
\hline
\textbf{Name} & \textbf{Original} & \textbf{After Down-sampling} & \textbf{Merging 4 Consecutive Records} & \textbf{Final Dataset} \\
\hline
DatasetFnusa -- Non seizure & $66581 \times 15000$ & $66581 \times 768$ & $66581/4 = 16645.25$ & $16645 \times 3072$ \\
\hline
DatasetFnusa -- Ictal & $18184 \times 15000$ & $18184 \times 768$ & $18184/4 = 4546.00$ & $4546 \times 3072$ \\
\hline
DatasetMayo -- Non seizure & $44259 \times 15000$ & $44259 \times 768$ & $44259/4 = 11064.75$ & $11064 \times 3072$ \\
\hline
DatasetMayo -- Ictal & $6099 \times 15000$ & $6099 \times 768$ & $6099/4 = 1524.75$ & $1524 \times 3072$ \\
\hline
\end{tabular}
}
\end{table}

\subsubsection{CHB-MIT dataset}

Among commonly available public datasets in EEG-based seizure detection research, the CHB-MIT dataset is widely recognized and used. The CHB-MIT dataset involves long-term scalp EEG recordings collected from 23 pediatric patients aged from 1.5 years to 22 years. EEG signals were recorded from 23-28 channels at a sampling frequency of 256 Hz. The dataset comprises around 500 hours of continuous EEG recordings. 173 seizures, annotated by neurologists, are present in the dataset. Given its detailed seizure annotations and wide usage in epilepsy research, the CHB-MIT dataset can be used as a benchmark dataset to test the seizure detection algorithm.

\begin{table}[h!]
\centering
\caption{Dataset Details of the EEG-iEEG Combined Dataset}
\label{tab:eeg_ieeg_combined}
\begin{tabular}{|l|c|c|}
\hline
\textbf{Final Combined Dataset} & \textbf{Records} & \textbf{Total} \\
\hline
Non-seizure (iEEG only) & 27,709 & 27,709 \\
\hline
Ictal (iEEG + CHB-MIT) & 6,070 + 885 & 6,925 \\
\hline
Preictal (CHB-MIT only) & 8,613 & 8,613 \\
\hline
\end{tabular}
\end{table}

\subsubsection{Multicentre intracranial EEG dataset}

The Multicentre iEEG data were used to represent intracranial EEG signals. The iEEG data is freely available from Kaggle. It contained two folders, namely DatasetFnusa and DatasetMayo. Each iEEG dataset folder contains four categories of folders: power-line noise (60Hz), artifacts, physiological/non-seizure activity, and pathological/seizure activity. Out of these categories, the physiological group includes brain activity that does not correspond to a seizure, while the pathological group includes activity that is seizure-related. The iEEG signals have been sampled at 5000 Hz, and each event has a length of 3 seconds of recording, which corresponds to 15000 samples per event. The temporal detail, in combination with intracranial recordings, makes the iEEG dataset appropriate for studies of seizure-related neuronal dynamics.

\subsubsection{Multimodal EEG–iEEG Dataset Creation}

To construct the suggested multimodal seizure dataset, part of the recordings from the CHB-MIT EEG database and the Multicentre iEEG database are selected. From the Multicentre Clinic iEEG database, we take Physiological (Non- seizure) and Pathological (Ictal) segments. For the CHB-MIT EEG database, we take only Preictal and Ictal segments and reject the non-seizure recordings. Hence, our resulting multimodal dataset contains 3 classes relevant in clinical settings: Non-Seizure, Preictal, and Ictal; and it was generated from overlapping EEG and iEEG signals. This approach leverages the common information provided by the two types of signals while maintaining the characteristics of the seizure status. As the EEG and iEEG signals are characterized and recorded differently, they cannot be straightforwardly combined. Thus, a Unified multimodal signal harmonization (UMSH) method is proposed to normalize both modalities and transform them to a common feature space for uniform seizure classification.

\subsection{Unified Multimodel Signal Harmonization}

The proposed UMSH framework proposed here consists of two main stages, iEEG Signal Harmonization and EEG Signal Harmonization, for the two modalities to standardize the heterogeneous natures of both and provide a uniform representation for multimodal seizure classification.

\subsubsection{iEEG Signal Harmonization}

The multicentre Clinic iEEG dataset is a set of 3-second-length signals with a sample frequency of 5000 Hz, whereas the CHB-MIT EEG dataset is sampled with 256 Hz. It would be impossible to fuse both datasets directly, as the temporal resolution difference is massive. First, the signals from the iEEG dataset were transformed to be compatible with EEG data. The original 5000 Hz signals were resampled to 256 Hz with Resample Poly (upsampling 32 and downsampling 625) to maintain the signal characteristics. The mathematical representation of resample poly is:

\begin{equation}
y(n)=\sum_{k=-\infty}^{\infty} x[k]\,h[nM-kL]
\label{eq:resample_poly}
\end{equation}

where $x[k]$ represents the input signal, $y(n)$ denotes the resampled output signal, $L$ is the upsampling factor, $M$ is the downsampling factor, and $h[n]$ represents the low-pass FIR filter used in polyphase filtering.
This transformed a 3-second segment of 15,000 samples to 768 samples. The duration of these short segments may not be sufficient to capture the entire dynamics of a seizure; the segments of the same class are concatenated for four subsequent time segments, creating a 12-second recording. Hence, each aligned iEEG sample has 3072 data points. This process is performed for both seizure and non-seizure classes.

\subsubsection{EEG Signal Harmonization}

The CHB-MIT dataset contains multichannel EEG data that is sampled at 256 Hz. We chose three clinically relevant EEG channels to create a concise representation consistent with iEEG data.  A single representative signal is produced by averaging the chosen channels. For seizure analysis, ictal segments are extracted from the labelled seizure intervals, and preictal segments from 10 minutes before seizure onset. All recordings were then split into non-overlapping 12-second windows. Because the sampling rate of the EEG was 256 Hz, all segments consisted of 3072 samples, which corresponds to the standard iEEG representation.

\subsubsection{Unified Multimodal Signal Harmonization}

EEG and iEEG signals have the same temporal properties after preprocessing, i.e., sampling frequency of 256 Hz, time duration of 12 seconds, and length of each segment as 3072. This will eliminate intra-modality variance and allow multiple modalities of brain signals to be unified into one feature space. The final data set contains four classes: EEG Preictal, EEG Ictal, iEEG non-Seizure, and iEEG Seizure. The produced combined data set is then used for class balancing and seizure classification. The pseudocode of UMSH given in Algorithm \ref{alg:umsh}.

\begin{figure}[t]
\centering
\includegraphics[width=0.95\columnwidth]{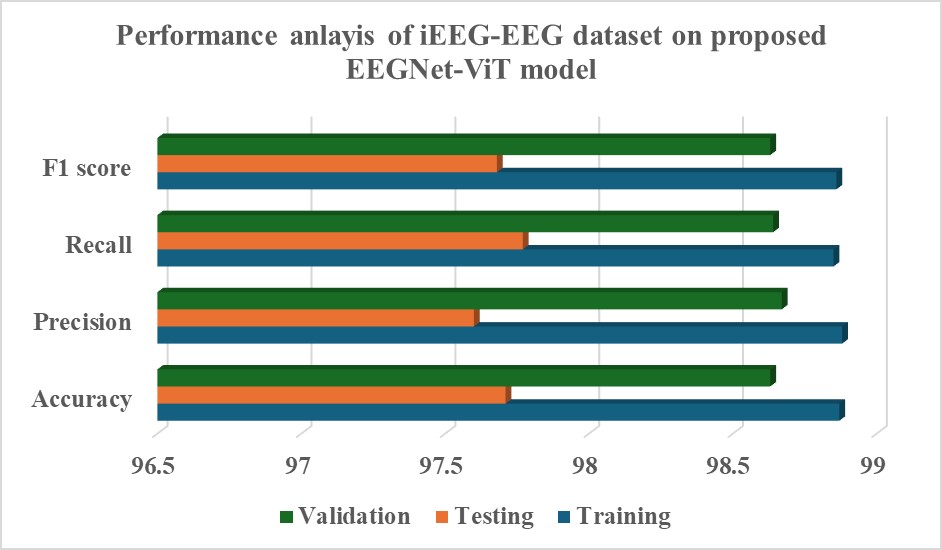}
\caption{Performance analysis of proposed EEG–iEEG Seizure Classification Framework on Training, Testing, and Validation Datasets.}
\label{fig:figure4}
\end{figure}

\begin{figure}[t]
\centering
\includegraphics[width=0.95\columnwidth]{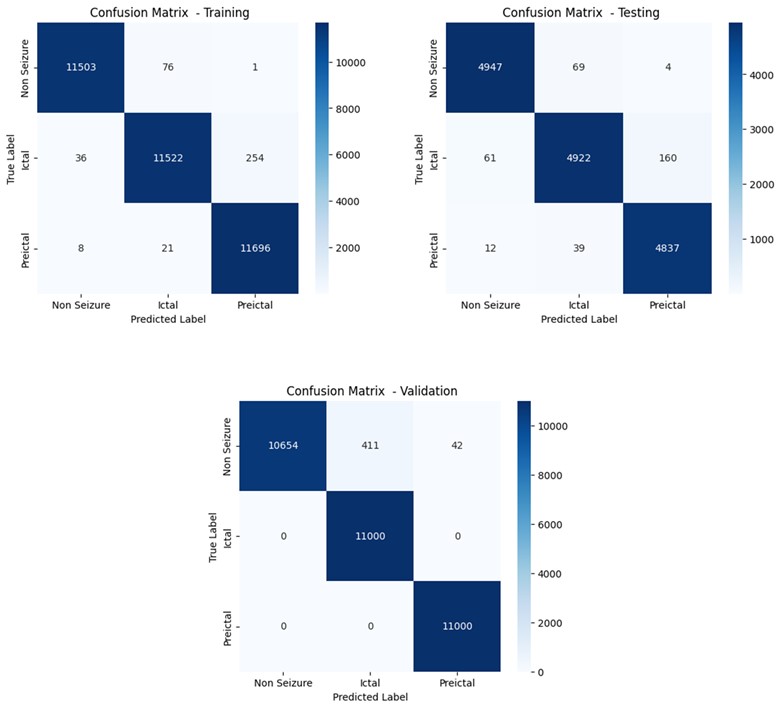}
\caption{Confusion matrix of proposed EEG–iEEG seizure classification model's performance across training, testing, and validation datasets.}
\label{fig:figure5}
\end{figure}

\begin{figure}[t]
\centering
\includegraphics[width=0.95\columnwidth]{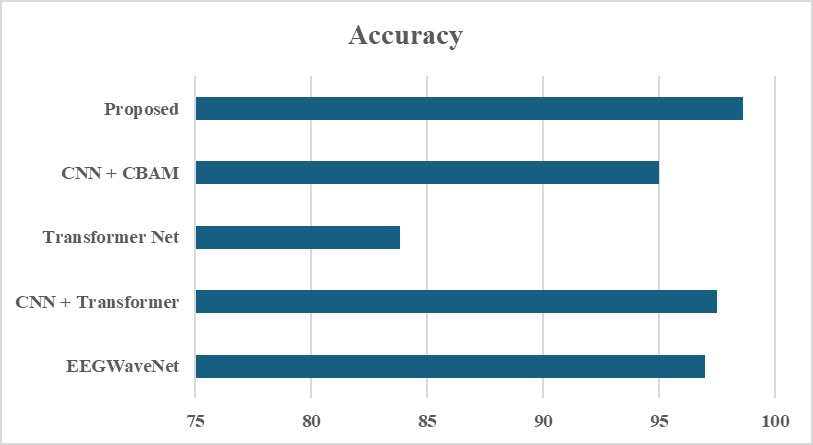}
\caption{Performance Comparison of the Proposed Framework with Existing State-of-the-Art Methods.}
\label{fig:figure6}
\end{figure}

The Unified Multimodal Signal Harmonization (UMSH) proposed framework, which integrates EEG and iEEG heterogeneous signals into one unique representation for seizure classification, is represented in Figure \ref{fig:figure2}. This framework comprises two fundamental stages: iEEG signal harmonization and EEG signal harmonization. Within the iEEG harmonization stage, multicentre iEEG recordings were first resampled from 5000 Hz to 256 Hz, using an upsampling factor of 32 and a downsampling factor of 625, then concatenated in 12-second intervals through the concatenation of 4 contiguous 3-second records, to keep the temporal coherence of data. Within the EEG signal harmonization stage, ictal and preictal EEG signals from CHB-MIT were selected, then concatenated into 12-second windows. Thus, both EEG and iEEG signals have the same duration and frequency. Then, data are forwarded to the standardization stage in which, using techniques presented below, differences among the two modalities are reduced, and data are then presented in a unique format. The standardized EEG and iEEG windows were used to form a multimodal dataset that is then used for seizure classification, which is then fed to our learning architecture. Table \ref{tab:ieeg_dataset_details} represents the dataset details of the iEEG dataset. From the iEEG dataset, we got 27709 records from the non-seizure category and 6070 records from the ictal category. Likewise, for the CHB-MIT dataset, we got 885 records for the ictal category and 8613 records for the preictal category.

Table \ref{tab:eeg_ieeg_combined} presents the details of the final combined EEG–iEEG dataset after preprocessing. The Non-Seizure class contains 27,709 samples while the Ictal and Preictal classes contain only 6,925 and 8,613 samples, respectively. Such class imbalance may result in favouring the majority class, thereby negatively impacting the performance in detecting seizure classes (which represent minority classes). As such, a relevant class balancing technique needs to be implemented in order to alleviate the class imbalance and enable the learning of consistent and general patterns of all classes. 

\subsection{Conditional Generative Adversarial Network (CGAN)-Based Data Balancing}

As is clear from Table III, the class distribution in the joint EEG-iEEG dataset is highly imbalanced, with a significantly larger number of samples in the non-Seizure class than the Ictal and Preictal classes. This imbalance in distribution may cause the learner to be biased towards the majority class, and lead to a low detection performance with bad generalization abilities to the seizure-related classes. It is necessary to apply a suitable data balancing method to equalize this class distribution before the classification phase. A Conditional Generative Adversarial Network (CGAN) is utilized to produce synthetic samples for the minority classes as a solution to the problem described. Instead of standard oversampling techniques, the CGAN produces class-specific synthetic samples by conditioning the network using the class labels in the training. Thus, the CGAN learns the distribution of each class and generates authentic synthetic EEG and iEEG samples and preserves characteristics of each class.

The CGAN framework includes two adversarial networks: the generator (G) and the discriminator (D). The generator takes as input the noise vector z and class label y and generates a fake sample, whereas the discriminator takes as input the combination of real and generated samples along with their class labels, and tries to discriminate between real and fake samples. By using adversarial learning, the generator can learn to generate better fake minority class samples. The objective function of CGAN is:

\begin{equation}
\begin{split}
\min_{G}\max_{D}V(D,G)=
\mathbb{E}_{x\sim p_{data}(x)}[\log D(x|y)] \\
+ \mathbb{E}_{z\sim p_{z}(z)}[\log(1-D(G(z|y)))]
\end{split}
\tag{2}
\label{eq:cgan_objective}
\end{equation}

where $x$ represents the real EEG/iEEG data sample, $z$ denotes the latent noise vector, and $y$ represents the class label. Here, $p_{data}(x)$ denotes the real data distribution and $p_z(z)$ represents the latent noise distribution. The generator $G$ learns to generate realistic class-conditioned samples, while the discriminator $D$ attempts to distinguish real samples from synthetic samples.

In the proposed framework, we employ CGAN to over-sample the minority classes, i.e., Ictal and Preictal classes, until a balanced class distribution is obtained. By introducing artificial samples, we enhance the data variability and decrease the data skew due to class imbalance. As a result, the class-balanced data allows the subsequent classifier to learn more distinctive seizure-related representations, which ultimately contribute to improving the robustness and generalization ability of the classification model. The resulting class-balanced EEG-iEEG data from CGAN over-sampling are used to train our proposed unified seizure classification framework. For every case, we extracted 10000 synthetic samples. 

\begin{table}[h!]
\centering
\caption{EEGNet-Transformer Layer Configuration}
\label{tab:eegnet_transformer_summary}
\resizebox{\linewidth}{!}{
\begin{tabular}{p{5.5cm}cc}
\hline
\textbf{Layer} & \textbf{Output Shape} & \textbf{Parameters} \\
\hline

InputLayer & (None, 3072, 1, 1) & 0 \\

Conv2D (F1, kernel\_size = (64,1)) & (None, 3072, 1, 16) & 1024 \\

BatchNormalization & (None, 3072, 1, 16) & 64 \\

Activation (ELU) & (None, 3072, 1, 16) & 0 \\

DepthwiseConv2D (kernel\_size = (1,16), depth\_multiplier = D) & (None, 3072, 1, 16) & 16 \\

BatchNormalization & (None, 3072, 1, 16) & 64 \\

Activation (ELU) & (None, 3072, 1, 16) & 0 \\

AveragePooling2D (pool\_size = (4,1)) & (None, 768, 1, 16) & 0 \\

Dropout (drate) & (None, 768, 1, 16) & 0 \\

SeparableConv2D (F2, kernel\_size = (16,1)) & (None, 768, 1, 32) & 768 \\

BatchNormalization & (None, 768, 1, 32) & 128 \\

Activation (ELU) & (None, 768, 1, 32) & 0 \\

AveragePooling2D (pool\_size = (8,1)) & (None, 96, 1, 32) & 0 \\

Dropout (drate) & (None, 96, 1, 32) & 0 \\

Lambda & (None, 96, 32) & 0 \\

TimeDistributed(Dense(embed\_dim)) & (None, 96, 64) & 2112 \\

PositionalEmbedding(max\_len, embed\_dim) & (None, 96, 64) & 6272 \\

TransformerBlock 1 (embed\_dim, num\_heads, mlp\_dim, drateT) & (None, 96, 64) & 33472 \\

TransformerBlock 2 (embed\_dim, num\_heads, mlp\_dim, drateT) & (None, 96, 64) & 33472 \\

GlobalAveragePooling1D & (None, 64) & 0 \\

LayerNormalization ($\epsilon=1e^{-6}$) & (None, 64) & 128 \\

Dropout (0.3) & (None, 64) & 0 \\

Dense (3) & (None, 3) & 195 \\

\hline
\end{tabular}
}
\end{table}

\subsection{Improved EEGNet-ViT architecture for multimodal classification}

The harmonized and class-balanced standard EEG and iEEG signals are given to the EEGNet architecture as input feature extraction after being normalized. Figure \ref{fig:figure3} represents the proposed EEGNet-ViT architecture. The reasons EEGNet was chosen as the first feature extraction module were its lightweight structure, computation-efficient design, and its ability to learn spatial-temporal features from electrophysiological signals effectively. While most standard CNNs require large amounts of trainable parameters, EEGNet utilized depth-wise and separable convolutions to learn features efficiently.

Input to the network is synchronized EEG/iEEG epochs with dimensions of 3072x11. First, a temporal convolution layer of F1=16 filters with kernel size (64,1) is applied to the input signal to extract frequency-specific characteristics of seizure/non-seizure activities, act as a temporal filter bank to extract patterns, and to exploit the temporal dependencies of brain signals. The temporal convolution expressed as:

\begin{equation}
F_t=\sigma(W_t * X+b_t)
\tag{3}
\label{eq:temporal_conv}
\end{equation}

where $W_t$ represents the temporal convolution kernel, $b_t$ denotes the bias term, $*$ represents the convolution operation, and $\sigma(\cdot)$ denotes the activation function. Following the temporal convolution layer, the extracted features are normalized with Batch Normalization and applied to the Exponential Linear Unit (ELU) activation function to provide the network the ability to learn non-linear features and to speed up convergence.

\begin{table}[h!]
\centering
\caption{Parameter Details of the Proposed Model}
\label{tab:model_parameters}
\begin{tabular}{|c|c|}
\hline
\textbf{Parameter} & \textbf{Value} \\
\hline
$F_1$ & 16 \\
$D$ & 2 \\
$F_2$ & 32 \\
$d_{rate}$ & 0.4 \\
$embed_{dim}$ & 64 \\
$num_{heads}$ & 4 \\
Depth & 3 \\
$mlp_{dim}$ & 128 \\
$d_{rateT}$ & 0.1 \\
$max_{tokens}$ & 98 \\
\hline
\end{tabular}
\end{table}

After time filtering, the first Depthwise Convolution layer is used with kernel size (1,16) and depth multiplier D=2. The depth wise convolution is expressed as:

\begin{equation}
F_s=\sigma(W_s \odot F_t+b_s)
\tag{4}
\label{eq:depthwise_conv}
\end{equation}

where $W_s$ denotes the spatial filter, $F_t$ represents the temporal feature map obtained from the previous layer, $b_s$ denotes the bias term, $\odot$ represents the depthwise convolution operation, and $\sigma(\cdot)$ denotes the activation function.

In this operation, the network learns spatial representations of the extracted temporal features. Compared to a normal convolution operation, it greatly reduces the number of trainable parameters. Batch Normalization and ELU activation are applied once more. Then, to reduce the number of feature dimensions and remove the redundant information, an Average Pooling layer with a pooling size (4,1) is applied, so that the dimension of the feature representation is decreased from 3072 samples to 768 samples in the temporal domain. After that, a dropout layer with a dropout rate of 0.4 is implemented to overcome overfitting and enhance model generalization performance.
Finally, the compressed feature maps are then fed to a Separable Convolution layer with F2=32 filters and the size (16,1). Separable convolution separately does the depth-wise and pointwise convolutions, thus making it computationally more efficient with preserving discriminative information from feature maps. This layer extracts and updates the temporal-spatial information that are learned from EEG and iEEG signals. These feature maps are normalized and activated with Batch Normalization and ELU activation functions. Subsequently, a second Average Pooling layer is introduced with a pooling size of (8,1), reducing the feature dimension from 768 to 96 temporal tokens. A dropout layer is then applied to enhance robustness and avoid overfitting of the model. The resultant feature dimension at this point is 9632, and the features have concise yet highly informative spatio-temporal features of the seizure. 

The extracted features are subsequently flattened into a sequential representation, where a \textit{Lambda} layer reshapes the feature maps, and a \textit{TimeDistributed Dense} layer projects each feature vector into a 64-dimensional embedding space (\texttt{embed\_dim = 64}). This embedding projection converts the EEGNet-derived feature maps into a compact token sequence while preserving the inherent temporal and spatial information. The generated token embeddings are then fed into the Vision Transformer (ViT) to capture global contextual information and long-range dependencies effectively. Table \ref{tab:eegnet_transformer_summary} illustrates the layer configuration of the EEGNet-ViT model, while Table \ref{tab:model_parameters} presents the detailed model parameters.

\section{Results and Analysis}

The efficacy of the proposed generalized seizure classification scheme was studied with the help of combined EEG-iEEG dataset. Extensive experiments were performed on various parameters to observe the effect of multimodal signal harmonization, class balancing using CGAN and EEGNet-Vision Transformer network in seizure classification accuracy. Many performance evaluation indices like accuracy, precision, recall, F1-score, sensitivity, specificity, AUC were used.

To determine the model performance and generalization, the combined EEG-iEEG data was split into two groups, namely training and validation set. To achieve balanced learning for all the seizure states, the data from non-seizure, ictal and preictal class were selected with care, both from the original and synthesized dataset. For the training section, we collected 16,600 samples of non-Seizure using 8,300 of each record of Dataset Fnusa and Dataset Mayo. We collected 6,070 original iEEG ictal samples from Dataset Fnusa, 885 ictal samples of CHB-MIT and generated 16,600 of synthetic seizure samples with CGAN to make Ictal set. In the same way, 8,613 original CHB-MIT samples together with 16,600 synthetic samples created with CGAN make Preictal set. For both ictal and preictal case, the synthetic generated samples were utilised for training and original samples utilised for validation. The training set have total number of samples as 49800, which is the sum of all three sets of classes as showed in Table \ref{tab:training_dataset}. During model training, the dataset was divided into 80\% for training and 20\% for testing.

\begin{table}[h!]
\centering
\scriptsize
\caption{Training Dataset Used for Model Development}
\label{tab:training_dataset}
\resizebox{\columnwidth}{!}{
\begin{tabular}{|p{2.5cm}|p{3cm}|c|c|}
\hline
\textbf{Class} & \textbf{Source} & \textbf{Total} & \textbf{Grand Total} \\
\hline

Non-seizure (iEEG) &
Original -- DatasetFnusa (First 8300) + DatasetMayo (First 8300) &
16600 & 16600 \\
\hline

Ictal (iEEG-Synthetic) &
CGAN Generated &
8300 & \multirow{2}{*}{83} \\
\cline{1-3}

Ictal (CHB-MIT-Synthetic) &
CGAN Generated &
8300 & \\
\hline

Preictal (CHB-MIT-Synthetic) &
CGAN Generated &
16600 & 16600 \\
\hline

\multicolumn{3}{|c|}{\textbf{Total}} & \textbf{49800} \\
\hline

\end{tabular}
}
\end{table}

To test model robustness and prevent data leakage, another dataset (validation dataset) was created from the available samples that did not take part in training. To do so, from the records that remained after excluding the training records and belonging to the non-Seizure class of Dataset Fnusa and Dataset Mayo, 7000 samples were chosen. For ictal and preictal case selected original samples. Therefore, the validation dataset comprised of 20955 samples as it can be observed in Table \ref{tab:validation_dataset}.

\begin{table}[h!]
\centering
\scriptsize
\caption{Validation Dataset Used for Model Evaluation}
\label{tab:validation_dataset}
\resizebox{\columnwidth}{!}{
\begin{tabular}{|p{2.5cm}|p{3cm}|c|c|}
\hline
\textbf{Class} & \textbf{Source} & \textbf{Total} & \textbf{Grand Total} \\
\hline

Non-seizure (iEEG) &
Remaining DatasetFnusa + DatasetMayo Samples &
8344 + 2763 & 11107 \\
\hline

Ictal (iEEG-Synthetic) &
CGAN Generated (Unused in Training) &
5500 & 11000 \\
\hline

Ictal (CHB-MIT-Synthetic) &
CGAN Generated (Unused in Training) &
5500 & \\
\hline

Preictal (CHB-MIT-Synthetic) &
CGAN Generated (Unused in Training) &
11000 & 11000 \\
\hline

\multicolumn{3}{|c|}{\textbf{Total}} & \textbf{33107} \\
\hline

\end{tabular}
}
\end{table}

The selection of such partitioning scheme can guarantee the proposed model to be trained with balanced multiclass data and simultaneously evaluated with a distinct validation set for the proposed EEG-iEEG seizure classification unified framework.
Figure \ref{fig:figure4} shows the classification accuracy, precision, recall, and F1-score of proposed EEG-iEEG seizure classification framework in training, testing and validation datasets. The accuracy of proposed framework in training, testing and validation set are 98.87\%, 97.71\% and 98.63\% respectively. In addition, with precision as 98.88\%, 97.60\% and 98.67\% it represents that the framework effectively classifies seizure related classes and the number of false positive class will be minimum. With the recall of 98.85\%, 97.77\% and 98.64\% it represents that framework correctly predict seizure class in various datasets. In addition, with F1-scores of 98.86\%, 97.68\% and 98.63\% confirms balanced precision and recall rate. The training, testing and validation set having good accuracy demonstrates that the proposed framework have good generalization property.      

\begin{table}[h!]
\centering
\caption{Ablation Study of Different EEG and iEEG Modality Configurations for Seizure Classification}
\label{tab:ablation_study}
\begin{tabular}{|l|c|}
\hline
\textbf{Modalities} & \textbf{Accuracy (\%)} \\
\hline
EEG Training, iEEG Testing & 95.45 \\
iEEG Training, EEG Testing & 94.67 \\
EEG + iEEG (Proposed) & 98.63 \\
\hline
\end{tabular}
\end{table}

The confusion matrix of proposed EEG–iEEG seizure classification model's performance across training, testing, and validation datasets is shown in Figure \ref{fig:figure5}. There are several correctly categorized samples for the Non-Seizure, Ictal, and Preictal classes in all three instances when the diagonal values are prevalent. There are very few misclassifications, mostly between the clinically comparable phases of Ictal and Preictal states. Testing findings demonstrate great generalization capabilities, and the training matrix demonstrates nearly perfect learning. Additionally, the validation matrix shows very low error rates and steady, reliable predictions. Overall, the model performs robustly and reliably in identifying seizure-related stages. Additionally, the performance and efficacy of the proposed framework were assessed by contrasting it with current state-of-the-art techniques.

Figure \ref{fig:figure6} shows the performance comparison of the proposed framework with existing state-of-the-art deep learning architectures. Among the existing architectures, CNN+Transformer \cite{sun2022continuous} resulted in maximum classification accuracy as 97.5\%, followed by EEGWaveNet \cite{thuwajit2021eegwavenet} (97\%), CNN+CBAM \cite{ouichka2022deeplearning} (95\%) and Transformer Net \cite{si2023lightweightcnn} (83.81\%). Proposed architecture achieved classification accuracy of 98.63\%, highest amongst all. Proposed framework achieved the highest gain over Transformer Net (1.63\%), CNN+CBAM (1.13\%), EEGWaveNet (3.63\%), and CNN+Transformer (14.82\%). The results indicate that proposed framework learns effective discriminative patterns of seizures from EEG-iEEG and the overall classification performance is significant than others.
\subsection{Ablation Analysis of EEG and iEEG Modalities}

To further investigate the impact of different configurations of input modality on seizure classification performance, an ablation study was performed on three settings; (i) training the model on EEG and testing on iEEG; (ii) training the model on iEEG and testing on EEG; and (iii) training and testing on both the EEG and iEEG signals.

As can be seen from Table \ref{tab:ablation_study}, the accuracy achieved in EEG Training-iEEG Testing setting was 95.45\%, and 94.67\% accuracy was gained in iEEG Training-EEG Testing setting. While cross-modal settings show the good generalisation capability, they perform worse than both EEG and iEEG joint setting. The best accuracy was achieved when the model was trained and tested on the EEG and iEEG signals jointly. Accuracy rate 98.63\% of EEG and iEEG signals combined shows that providing information from both modalities enables the model to learn the features associated with seizure effectively.
\subsection{Ablation Analysis of proposed framework}

Table \ref{tab:cgan_ablation} shows the performance of the proposed model in two experimental scenarios: without CGAN augmentation, and with UMSH and CGAN in the second setting. In the first setting, the samples of the original dataset were used for training with 30\% of the total data used for validation.
\begin{table}[h!]
\centering
\caption{Effect of CGAN-based augmentation on seizure classification performance}
\label{tab:cgan_ablation}
\resizebox{\columnwidth}{!}{%
\begin{tabular}{|l|c|c|c|c|}
\hline
\textbf{Configuration} & \textbf{Accuracy (\%)} &
\textbf{Precision (\%)} & \textbf{Recall (\%)} &
\textbf{F1 Score (\%)} \\
\hline
Without CGAN & 94.76 & 94.77 & 94.76 & 94.76 \\
\hline
With UMSH and CGAN & 98.18 & 97.93 & 98.18 & 98.05 \\
\hline
\end{tabular}%
}
\end{table}

\section{Conclusion}
In this paper, we introduced a new framework to classify seizure based on multimodal EEG and iEEG signals, thereby to overcome the shortcomings of single modality seizure detection systems. Developed a UMSH framework to normalize heterogeneous EEG/iEEG data recorded with different sampling rates and properties. To overcome the class imbalance issue, a cGAN-based data augmentation was applied for diversifying the number of classes involved. Besides that, a hybrid EEGNet-Vision Transformer (EEGNet-ViT) model was used for learning local and global patterns associated with seizure. According to the experimental results, the proposed framework was successful to classify seizures robustly. Also, the ablation study revealed the merit of the multimodal data fusion approach. For EEG Training-iEEG Testing and iEEG Training-EEG Testing setup, the classification accuracy was found to be 95.45\% and 94.67\% respectively. But the model trained and tested using both EEG and iEEG signals gave the highest accuracy of 98.63\%. This result showed that the multimodal EEG and iEEG signals could represent more information related to seizure, thus facilitating the model to learn more sophisticated and discriminative representations compared to single-modality settings. The proposed method provided superiority to current methods and increased classification accuracy and reliability. It proved the feasibility of joint EEG-iEEG learning for future epilepsy monitoring and clinical decision making, indicating a favourable prospect in future multimodal seizure analysis studies. Future research will concentrate on expanding the proposed seizure prediction framework, adding more physiological modalities, introducing explainable AI methods for better interpretability, and validating the model on bigger multicenter datasets.

\bibliographystyle{IEEEtran}
\bibliography{references.bib}

\end{document}